\documentclass[12pt]{article}

\usepackage{amssymb}
\usepackage{amsfonts}

\begin{document}

\title {Genome as a functional program}

\author{S.V. Kozyrev\footnote{Steklov Mathematical Institute of Russian Academy of Sciences,
ul. Gubkina 8, Moscow 119991, Russia}}

\maketitle

\begin{abstract}
We discuss a model of genome as a program with functional architecture and consider the approach to Darwinian evolution as a learning problem for functional programming. In particular we introduce a model of learning for some class of functional programs. This approach is related to information geometry (the learning model uses some kind of distance in the information space).
\end{abstract}






\section{Introduction}

In the present paper we discuss the problem  --- which mathematical object could be relevant for description of work of a genome and corresponding set of chemical reactions in a cell? Popular approach is to use networks (graphs), in particular metabolic network  (network of chemical reactions in a cell, edges correspond to reactions and vertices (nodes) correspond to reagents and reaction products), another example is a network of interacting genes (vertices are genes encoding molecules, edges connect molecules which interact in some sense). It was mentioned that these graphs in many cases are scale free networks, i.e. graphs where the number of edges incident to a vertex is distributed according to power law. This kind of scaling is also observed for sizes of families of paralogous genes (genes in a genome generated by duplication events). To describe these phenomena E.V.Koonin proposed to consider a genome as a ''gas of interacting genes'' \cite{Koonin1}, the scaling in this approach should be explained by Gibbs distribution for this model. Y.I.Manin for explanation of the Zipf's scaling law of frequencies of words in texts proposed a model of statistical mechanics with the Hamiltonian given by Kolmogorov complexity \cite{Manin} (''complexity as energy''). In \cite{Scaling} these two approaches were discussed in relation to biological evolution as model of learning.

Sometimes genomes are discussed using a metaphor of program. If we take this metaphor seriously, we have two questions --- how this program can operate and how this program can be modified by biological evolution? It seems random modifications of a working program should ruin a program completely. The operation of the program is also mysterious --- a genome is a set of genes which work in parallel, i.e. a genome is a massively parallel program which contains a set of parallel processes correspondent to genes. Number of genes in viruses could vary enormously (usually dozens), in bacteria the number of genes can be estimated by several thousands and in multicellular organisms by several dozens of thousands. Genes are subjected to regulation of expression and majority of genes usually is switched off but it is obvious that genome as a program is massively parallel.

Provision of high degree of parallelism is one of major problems in programming. One of important approaches to this problem is given by functional programming based on lambda calculus. In functional programs, unlike in imperative programs, execution of a program is related not to sequential modification of states, but to application of functions. Rejection of states (or considerable simplification of states) allows to eliminate errors related to access of different parts of a program to the same states. One of general methods of functional programming is recursion, or self--reference. Parallelism of computations in functional programs is provided by the Church--Rosser property.

Biological evolution from point of view of programming reduces to generation of a program given data (evolution by selection pressure), i.e. to a problem of machine learning. Let us note that learning theory for functional programs is not yet developed.

Above discussion leads to following conclusions:

1) Genome is a functional program based on parallel execution of interacting agents, or genes (''interacting gas of genes'').

2) This program is recursive, genes can refer to results of applications of other genes, i.e. functional program for a genome is given by fixed point of a genome as a list of genes in the sense of lambda calculus, ''life is a fixed point of genome''.

3) Reduction graph of a functional program for genome is related to metabolic network for genome.

4) Biological evolution reduces to a problem of learning for functional programming.

We will propose a model of learning for some family of functional programs modeling genomes, this model of learning will be described by a statistical sum which contains summation over paths in the reduction graph of the program of Gibbs factors of some functional of action along the path. This action functional can be considered as estimate for Kolmogorov complexity of computation along the reduction path. From the point of view of learning theory the functional of action provides regularization of the model of learning (reduction of the effective complexity of the program), from the point of view of biology this functional describes effort needed to generate a set of molecules. Therefore the model under consideration combines properties of the physical model of ''interacting gas of genes'' and the model of ''complexity as energy''.

Approach of the present paper can be compared to information geometry approach, see \cite{Chentsov}, \cite{Amari}, \cite{Marcolli}. We consider model of statistical mechanics in information geometry with the functional of action related to estimate for Kolmogorov complexity for some functional program.

Exposition of this paper is as follows.
In Section 2 we describe the construction of genome as a functional program and introduce learning model for functional programs. In Section 3 (Appendix 1) we discuss relation to grammars by Chomsky and in Section 4 (Appendix 2) we discuss the analogy to alignment of sequences --- editing operations used in alignment give a simplest example of operations discussed in Sections 2, 3 and the score functional optimized in alignment gives example of the action functional considered in Section 2.

\section{Genome as a program}

We use the notations by John Backus \cite{Backus} (systems of functional programming, or FP systems). FP systems operate with functions which map objects to objects. Functional forms (in particular composition of functions operation $\circ$) allow to construct new functions using earlier defined functions.

In this model the set of objects is the set $S$ of sets of finite words with multiplicity (any word in object has multiplicity given by a natural number, any object contains finite number of words with non-zero multiplicity). The program is recursively defined by a list of functions $G=[g_1,\dots,g_n]$, any function $g_k$ is a map $S\to S$ (domain of the map can be smaller than $S$).

Moreover this map can multivalued (in lambda calculus this means that application of function $g_k$ to object $v$ is represented by a lambda term where reduction can be done in several ways). For example, function $g_k$ could perform the operation of cleavage of a word $v$ in two words in the position of some special sub-word $w$, word $v$ can contain several sub-words $w$ and cleavage of $v$ can be performed at position of any of these sub-words, also $g_k$ could act at sub-words $w$ belonging to different words in the object. List of functions $G=[g_1,\dots,g_n]$ can be considered as a multivalued map $G:S\to S$ where we can apply to object $v\in S$ any function $g_k$ in the list $G$ (recall the function $g_k$ itself is multivalued). List of functions $G$ is related to some generalization of context free grammar by Chomsky \cite{Chomsky}, see Appendix 1.

We will consider a family of functional programs defined recursively by a list of functions $G=[g_1,\dots,g_n]$
\begin{equation}\label{Genome}
\widetilde{G} = \widetilde{G} \circ G = [\widetilde{G}\circ g_1,\dots,\widetilde{G}\circ g_n].
\end{equation}

Here $\circ$ is the functional form of composition, $n$ is a number of genes in a genome. The list of functions $G$ (the genome) is given and the above equation  defines the functional program $\widetilde{G}$ recursively as a map $S\to S$,  i.e. the program $\widetilde{G}$ is a fixed point of the genome $G$ as a lambda term.

Interpretation: set $S$ of lists of words with multiplicity is the set of combinations of biological sequences (molecules), $g_k$ is a gene encoding protein which performs transformation of biological sequences  (chemical reaction). Moreover any gene is represented by some biological sequence, i.e. genome $G$ lies in $S$.

Let $v_0\in S$ be some set of words with multiplicity. Let us construct the graph $\Gamma_{\widetilde{G}}(v_0)$ (in interpretation by lambda--calculus this corresponds to reduction graph of the program $\widetilde{G}$ in ''lazy'' evaluation strategy). Vertices of the graph correspond to some objects in $S$, edges correspond to applications of functions $g_k$ (genes), the graph is constructed as follows. Let us apply $\widetilde{G}$ to $v_0$, by recursive definition (\ref{Genome}) any $g_k$ in the genome $G$ can be applied to $v_0$ (in non-unique way, as discussed above), this gives a set of vertices of the graph. At the next step we apply to obtained vertices all $g_j\in G$. We include to graph $\Gamma_{\widetilde{G}}(v_0)$ all vertices and edges obtained by iteration of this procedure (objects in $S$ can be obtained in non-unique way, we identify vertices which coincide as sets of words with multiplicity). This graph can be interpreted as the metabolic network for the genome $G=[g_1,\dots,g_n]$.

\medskip

\noindent{\bf Remark}. The program $\widetilde{G}$ defined by (\ref{Genome}) is a multivalued map $S\to S$, application of $\widetilde{G}$ to $v_0\in S$ gives the set of vertices of the graph $\Gamma_{\widetilde{G}}(v_0)$. This point of view is non-standard --- in the standard approach the result of execution of a functional program is given by the normal form of the corresponding lambda term (if this form exists).

\medskip

Let us put in correspondence to action of a function $g_k$ a positive weight $K(g_k)$, and to oriented path $p$ from $v_0$ to $v$ in graph $\Gamma_{\widetilde{G}}(v_0)$ we put in correspondence the functional of action equal to the sum of weights of edges in this path
\begin{equation}\label{Action}
K_{\widetilde{G}}(p)=\sum_{k}K(g_{i_k}).
\end{equation}
This functional can be considered as the cost of computation along the path $p$ or weighted estimate for Kolmogorov complexity of generation of $v$ starting from $v_0$.

\medskip

\noindent{\bf Temperature learning}.
Problem of machine learning is the minimization over the space of parameters $s$ of the sum of the loss functional and the regularization term
$$
H(s)=R(s)+Reg(s)\to\min.
$$

Definitions of the loss functional $R$ and regularization $Reg$ depend on particular problem of machine learning. Here we will use action functional (\ref{Action}) as a regularization.

Regularization is important for control of overfitting (to decrease the entropy of the space of parameters $s$). Learning at non-zero temperature means that instead of minimization we consider the statistical sum over values of the parameter $s$, $\beta>0$ is the inverse temperature
$$
Z=\sum_{s}e^{-\beta H(s)}.
$$

\medskip

\noindent{\bf Temperature learning for functional programs}. Let us consider a real valued function $F(v)$ on $S$ (the fitness function) and the statistical sum for the functional program (\ref{Genome})
\begin{equation}\label{Z[G,v_0]}
Z[\widetilde{G},v_0]=\sum_{v\in \Gamma_{\widetilde{G}}(v_0)}e^{-\beta F(v)}\sum_{p\in {\rm Path}\left(\Gamma_{\widetilde{G}}(v_0)\right):v_0\to v}e^{-\beta K_{\widetilde{G}}(p)},
\end{equation}
here summation over $p$ runs over all reduction paths in the graph $\Gamma_{\widetilde{G}}(v_0)$ with beginning in $v_0$ and end in $v$, $\beta$ is the inverse temperature. This statistical sum estimates fitness of the functional program $\widetilde{G}$ (in the Gibbs factor containing $F(v)$) and effort needed to generate $v$ (in the Gibbs factor containing the action functional $K_{\widetilde{G}}(p)$), summation over paths $p$ describes possibility to generate $v$ using different order of application of genes. Varying values $K(g_k)$ we will modify contributions to the statistical sum (\ref{Z[G,v_0]}) from different metabolic paths which describes gene regulation.

Genes in a genome $G=[g_1,\dots,g_n]$ (which were considered above as maps) are represented by biological sequences, i.e. genome $G$ itself can be considered as object in $S$.
This allows to describe biological evolution (transformation of a genome) by action of some ''evolutionary program'' $\widetilde{E}$ with ''evolution genes'' $E=[e_1,\dots,e_m]$ (in analogy to program (\ref{Genome}), evolution transforms genomes as lists of words to genomes)
\begin{equation}\label{Evolution}
\widetilde{E}= \widetilde{E} \circ E = [\widetilde{E}\circ e_1,\dots,\widetilde{E}\circ e_m].
\end{equation}

Evolution transforms genomes to genomes, this results in transformation of maps $g_k$ in (\ref{Genome}), scores $K(g_k)$, action functional (\ref{Action}) and transformation of statistical sum (\ref{Z[G,v_0]}).

Darwinian evolution is described as a learning problem at non-zero temperature with the statistical sum
\begin{equation}\label{Z[E,G_0]}
Z[\widetilde{E},G_0]=\sum_{G\in \Gamma_{\widetilde{E}}(G_0)}Z[\widetilde{G},v_0]\sum_{s\in {\rm Path}\left(\Gamma_{\widetilde{E}}(G_0)\right):G_0\to G} e^{-\beta' K_{\widetilde{E}}(s)}.
\end{equation}
Here we consider the graph $\Gamma_{\widetilde{E}}(G_0)$ of the evolution program defined as above, the summation runs over paths $s$ with beginning in the ancestor genome $G_0$ and end in the descendant genome $G$, then we sum over descendants $G$ in order to describe temperature learning with inverse ''evolution temperature'' $\beta'$.

Summation over paths describes parallelism of metabolic pathways in a cell (for statistical sum (\ref{Z[G,v_0]})) and parallelism in evolution (for statistical sum (\ref{Z[E,G_0]})). Gibbs factor in (\ref{Z[G,v_0]}) of the action functional (\ref{Action}) for program (\ref{Genome}) constrains the complexity of computations of the program $\widetilde{G}$ which give contribution to the statistical sum, analogously, Gibbs factor in (\ref{Z[E,G_0]}) of the action functional for the evolution program (\ref{Evolution}) constrains the complexity of evolutionary transformations in the statistical sum (\ref{Z[E,G_0]}). Presence of both these factors gives regularization for the problem of temperature learning for the functional program (regularization by estimate of Kolmogorov complexity). Transition from (\ref{Z[G,v_0]}) to (\ref{Z[E,G_0]}) can be considered as analogue of the replica transform used in the physics of disordered systems \cite{MPV}.

\medskip

\noindent{\bf Remark}. In computations for programs (\ref{Genome}), (\ref{Evolution}) at any step of recursion there are several options to take reduction, for computations of statistical sums (\ref{Z[G,v_0]}), (\ref{Z[E,G_0]}) one has to perform all combinations of reductions. This situation can be compared with computation at non-deterministic Turing machine (or NTM), where at some steps of computation NTM should duplicate and perform two branches of computation. Iteration of duplications allows to perform brute force search. The program (\ref{Genome}) can be considered as a functional version of program for NTM. Brute force search for statistical sum (\ref{Z[G,v_0]}) can be realized using parallel chemical reactions in a cell, and brute force search for statistical sum (\ref{Z[E,G_0]}) of evolution can be a result of selection for ensemble of evolving cells. Fitness function $F$ should belong to the class ${\bf P}$ of computable in polynomial time functions, hence the problem of Darwinian evolution (computation of statistical sum (\ref{Z[E,G_0]})) should be related to the class ${\bf NP}$ of computable in polynomial time at NTM functions. Class ${\bf NP}$ contains characteristic functions, thus for computation of the statistical sum (\ref{Z[E,G_0]}) the definition of the class should be generalized. The zero temperature limit of this statistical sum (the problem of finding of the most fit genome) is a characteristic function, therefore the above temperature learning problem is related to generalization of the class ${\bf NP}$ to non-zero temperatures.

\medskip

\noindent{\bf Summary}. A model of genome as functional program (\ref{Genome}) is proposed, the program is defined recursively by a list of genes (''life is a fixed point of genome''). Operation of this program is described by statistical sum (\ref{Z[G,v_0]}) (statistical sum for ''interacting gas of genes''). The statistical sum contains summation over reduction paths of Gibbs factors of the action functional (\ref{Action}) (cost of computation along the reduction path, or estimate for Kolmogorov complexity, ''complexity as energy''). Biological evolution is described by the functional program (\ref{Evolution}), Darwinian evolution by selection is the temperature learning problem (\ref{Z[E,G_0]}) for the program of evolution.

\section{Appendix 1: Grammar}

Set of chemical reactions in a cell can be considered as analogue of generative grammar by Chomsky. We consider finite strings of symbols from some alphabet, transformations from the grammar act on finite sets of finite strings with multiplicity, i.e. on the set $S$ of taking values in natural numbers functions with finite support on the set of finite strings (multiset of strings). $S$ can also be understood as set of formal sums of finite strings with natural coefficients, only finite number of coefficients is non-zero. Grammar is defined by a finite family of generative rules (genes), any rule is defined as follows: the rule is applicable to a subset in $S$ containing strings with some substrings, the transformation acts locally (at position of substrings). In particular, the set of generative rules can contain:

Gluing (or junction) of two strings with specific ends
$$
u'u+vv'\mapsto u'uvv';
$$
Cleavage of strings at position of specific substring
$$
u'uvv' \mapsto u'u+vv';
$$
Substitution of specific substring by the other
$$
u'uu''\mapsto u'vu'';
$$
Deletion of substring (substitution by empty substring) between two specific substrings
$$
u'uwvv'\mapsto u'uvv';
$$
Insertion of substring between two specific substrings (substitution of empty substring by specific substring)
$$
u'uvv' \mapsto u'uwvv';
$$
Insertion of nonspecific substring between two specific substrings
$$
u'uvu''+w\mapsto u'wu'';
$$
Duplication of text between two specific substrings
$$
u'u w vu''\mapsto u'u ww vu''.
$$

This set of rules defined by specific substrings $u$, $v$ defines the grammar. Domain of generative rule $g$ contains elements $s\in S$ of the form $s=s_1+s_2$, where operation $g$ is applicable to $s_1$, then the transformation acts as $g:s_1+s_2\mapsto gs_1+s_2$. Since this expansion is non-unique, the result of action of $g$ is also non-unique. This set of rules can be considered as a generalization of a context free grammar by Chomsky \cite{Chomsky}.

\section{Appendix 2: Alignment of sequences}

Alignment of two sequences is a way to transform one sequence to the other by editing operations, namely by combinations of insertions, deletions and symbol substitutions. Scores of operations are positive numbers and score of a combination of operations is a sum of scores of operations in this combination. These operations can be considered as a particular case of mappings $g_k$ in (\ref{Genome}) and the score of alignment gives an example of the action functional (\ref{Action}). Alignments can be found by dynamic programming algorithms which minimize alignment score (i.e. action (\ref{Action}), for this particular set of operations the algorithms are simple).

The standard definition of alignment is as follows \cite{Pevzner}.
Let ${\cal A}$ be a $k$-letter alphabet, ${\cal A}'={\cal A}\bigcup \{-\}$ be expanded alphabet where $\{-\}$ is a space symbol.
Let $V$, $W$ be finite sequences of symbols in ${\cal A}$.

\medskip

{\bf Alignment} of two sequences $V=v_1\dots v_n$ and $W=w_1\dots w_m$ is a matrix with two lines of equal length $l\ge n,m$, the first line is a sequence $\widetilde{V}=\widetilde{v}_1\dots \widetilde{v}_l$ obtained from $V$ by insertion of $l-n$ spaces, the second line is a sequence $\widetilde{W}=\widetilde{w}_1\dots \widetilde{w}_l$ obtained from $W$ by insertion of $l-m$ spaces. Columns with two spaces are forbidden.

Columns with spaces in the first line are called insertions and columns with spaces in the second line are called deletions.
Columns with equal symbols in both lines are called matches and columns with two different symbols are called mismatches.

Score of a column is a real number (depends on symbols in the column). Score of alignment is a sum of scores of columns
$$
\delta(\widetilde{V},\widetilde{W})=\sum_{i=1}^{l}\delta(\widetilde{v}_i,\widetilde{w}_i).
$$

\medskip

\noindent{\bf Example}: Scores $\delta(x,x)=0$ for matches, $\delta(x,y)=\mu>0$ for mismatches and $\delta(x,-)=\delta(-,x)=\sigma>0$ for insertions and deletions:
$$
{\rm score~}({\rm alignment})=\mu \#({\rm mismatches})+\sigma \#({\rm indels})
$$
(lower score --- better alignment).

\medskip

Alignment ($\widetilde{V}$, $\widetilde{W}$) of sequences $V$, $W$ corresponds to a combination of editing operations of sequence $V$ which converts $V$ to $W$. Editing operations correspond to columns of the alignment matrix ($\widetilde{V}$, $\widetilde{W}$) and can be performed in arbitrary order. Insertion operation is the insertion of space in sequence $V$ (at the position corresponding to the column), deletion operation is the deletion of the corresponding symbol in $V$, mismatch operation is the substitution of a symbol in $V$ by the corresponding symbol in $W$.

\end{document}